\documentclass{article}

\usepackage{arxiv}

\usepackage[utf8]{inputenc}
\usepackage[T1]{fontenc}

\usepackage{hyperref}
\usepackage{url}

\usepackage{graphicx}
\usepackage{float}
\usepackage{tikz}
\usetikzlibrary{arrows.meta,positioning,fit,shapes.geometric}

\usepackage{pgfplots}
\pgfplotsset{compat=1.18}

\usepackage{booktabs}
\usepackage{array}
\usepackage{tabularx}
\usepackage{multirow}
\usepackage{longtable}

\usepackage{amsmath}
\usepackage{amssymb}
\usepackage{amsfonts}

\usepackage{enumitem}

\usepackage{caption}
\usepackage{subcaption}

\usepackage[numbers,sort&compress]{natbib}

\usepackage{setspace}
\usepackage{xcolor}

\hypersetup{
    colorlinks=true,
    linkcolor=blue,
    citecolor=blue,
    urlcolor=blue
}

\setlist[itemize]{leftmargin=*}
\setlist[enumerate]{leftmargin=*}

\newcolumntype{Y}{>{\raggedright\arraybackslash}X}
\newcolumntype{P}[1]{>{\raggedright\arraybackslash}p{#1}}

\title{Agentic Data Cleaning Without a Clean Reference: An Experimental Study of Capabilities and Trade-offs}

\author{
 Hadi Fadlallah \\
 Faculty of Arts and Sciences\\
 University of Sciences and Arts in Lebanon\\
 Beirut, Lebanon \\
 \texttt{ha.fadlallah@usal.edu.lb}\\
 ORCID: 0000-0003-1160-5980
}

\begin{document}

\maketitle

\begin{abstract}
Data cleaning without a trusted clean reference is challenging because unusual values may represent either genuine errors or valid observations. This paper studies how different agent capabilities affect reference-free data cleaning and proposes an evidence-grounded framework that combines structured context, profiling, LLM reasoning, executable checks, controlled evidence retrieval, source ranking, citation alignment, conservative repair, reversible scripts, and provenance logging. Seven configurations are evaluated across financial, clinical, and environmental-monitoring datasets using controlled synthetic corruption and original-data descriptive analysis, resulting in 126 completed runs. The evaluation includes two comparison baselines and a progressive LLM-based sequence that adds executable tools, evidence retrieval, evidence controls, and conservative repair. In the synthetic evaluation, the deterministic profiling baseline achieved the highest detection F1-score of 0.561. Among the LLM-based configurations, the full conservative configuration achieved the highest F1-score of 0.421, but no configuration performed best across all evaluation criteria. The source-ranked configurations achieved the lowest unsupported-rule rates, while decision-level citation alignment remained weak. The full conservative configuration produced no unsafe or unnecessary modifications, although these rates were already zero before the conservative policy was added, and it performed no direct repairs. Overall, the results show that additional capabilities introduce trade-offs among detection, repair, evidence grounding, conservative behaviour, reproducibility, and operational cost rather than producing consistent improvements. The study provides a structured framework and empirical methodology for evaluating these trade-offs in reference-free agentic data cleaning.
\end{abstract}


\section{Introduction}
\label{sec:introduction}

Data cleaning is essential in analytics, scientific computing, data management, and machine-learning pipelines, but errors are rarely purely syntactic. Missing values, invalid formats, implausible codes, temporal irregularities, duplicates, and unusual measurements must be interpreted relative to task, consumer, and context. Foundational data-quality research frames quality as multidimensional and dependent on data consumers rather than as a single correctness property \citep{wang1996beyond}. Ontological and methodological perspectives further show that data-quality dimensions must be interpreted with respect to the represented real-world entities and the intended assessment context \citep{wand1996anchoring,pipino2002data}. Broader data-quality methodologies likewise treat assessment and improvement as context-sensitive processes involving dimensions, data types, techniques, and organizational requirements \citep{batini2006dataquality}. Classical cleaning work shows that profiling, constraints, external information, and human judgment are needed because no single technique detects or repairs all quality problems \citep{rahm2000cleaning}. Subsequent surveys and systems reinforce this view by highlighting the complementary roles of error detection, profiling, rule discovery, repair, and user involvement \citep{abedjan2016detecting,ilyas2019datacleaning}.

Large language models (LLMs) and agentic systems can infer column semantics, propose rules, generate scripts, explain anomalies, and coordinate multi-step workflows. Recent work has explored LLM-supported cleaning workflows and automated data-cleaning pipelines \citep{li2025autodcworkflow,biester2024llmclean}. Other studies examine retrieval-supported repair and agentic cleaning workflows that combine reasoning with external evidence or actions \citep{naeem2024retclean,qi2024cleanagent}. Related work also investigates contextual rule generation, semantic table profiling, and LLM-assisted data standardisation \citep{huang2024cocoon,zhang2024datacleaningllm}. These capabilities are promising, but they do not solve the central reference-free problem: a suspicious value may be a true error, a rare but valid signal, an undocumented convention, or a case requiring expert review. In domains such as clinical records, finance, and radiation monitoring, unsupported modification can damage scientific, operational, or regulatory meaning.

This paper therefore studies reference-free cleaning as evidence-grounded decision-making rather than as direct transformation. The proposed framework asks an agent to detect candidate issues, gather internal and external evidence, construct requirements, classify repair risk, generate reversible transformations only when justified, and preserve or escalate ambiguous cases. The study evaluates this design using two comparison baselines, A0 and A1, followed by a progressive ablation from A2 to A6. The evaluation covers three datasets, two evaluation settings, and three repeated runs, producing 126 successful runs. The aim is to measure configuration trade-offs under the tested datasets and controlled evidence setting, not to claim that one configuration is universally superior. Controlled synthetic-corruption runs provide ground truth for detection, repair, and valid-signal preservation, while original-data runs are interpreted descriptively because no trusted clean reference or manual adjudication is available.

The main contribution is threefold. First, the paper formalises the contextual knowledge, tool capabilities, evidence sources, and decision classes that may support reference-free data cleaning. Second, it proposes an evidence-grounded framework that combines profiling, tools, controlled evidence retrieval, source ranking, citation alignment, conservative repair, reversibility, and provenance. Third, it reports an exploratory ablation study that measures how these capabilities affect cleaning behavior. The results show clear trade-offs: A1 achieved the highest synthetic detection F1-score, A5 and A6 had the lowest unsupported-rule rates, and A6 showed more conservative repair behavior. However, the most complete configuration was not consistently better in detection, repair, reproducibility, runtime, or token cost.

The study is intentionally cautious in its claims. It does not assume that more agent capabilities necessarily produce better cleaning, and it does not evaluate original-data outputs as if they had hidden ground truth. Instead, it asks which components change observable behavior under controlled conditions and which uncertainties remain when the reference-free assumption is respected.

The paper is organised as follows. Section~\ref{sec:background} positions the study in relation to data-quality, cleaning, retrieval, provenance, and LLM-agent research. Sections~\ref{sec:problem} and~\ref{sec:taxonomies} define the reference-free setting and the taxonomy of contextual knowledge, tools, evidence, and decision classes. Sections~\ref{sec:framework}--\ref{sec:evaluation} present the framework, implementation, and metrics. Sections~\ref{sec:results} and~\ref{sec:discussion} report and interpret the ablation results, followed by threats to validity and the conclusion.

\section{Background and Related Work}
\label{sec:background}

Research on data quality provides the conceptual basis for this study. Quality dimensions such as accuracy, completeness, consistency, timeliness, believability, and interpretability depend on the consumer, intended use, and represented real-world entities \citep{wang1996beyond,wand1996anchoring,pipino2002data}. Broader methodologies organise assessment and improvement across dimensions, data types, techniques, and systems \citep{batini2006dataquality,batini2009methodologies}. Context-aware data-quality work further argues that assessment rules must be selected with respect to usage context and domain constraints \citep{fadlallah2023context,serra2024context}. The CTXDQ model provides a computational interpretation of this view by representing contextual characteristics and matching them to relevant quality dimensions and assessment techniques \citep{fadlallah2023ctxdq}. These ideas motivate the explicit cleaning context used in this paper.

Classical cleaning and repair systems show why reference-free cleaning cannot rely on natural-language reasoning alone. Profiling identifies distributions, types, missingness, duplicates, and dependencies \citep{naumann2014profiling}; duplicate and entity-resolution methods require specialised matching logic \citep{christen2012data}; and anomaly detection treats unusual observations as deviations from expectations, not automatically as errors \citep{chandola2009anomaly}. Systems such as Potter's Wheel and Wrangler emphasise interactive transformation and user feedback \citep{raman2001potters,kandel2011wrangler}, while NADEEF and constraint-based repair work formalise rule-driven detection and repair \citep{ebaid2013nadeef,bohannon2005cost,fan2012foundations}. Probabilistic systems such as HoloClean and PClean incorporate uncertainty, signals, and domain knowledge \citep{rekatsinas2017holoclean,lew2021pclean}. ActiveClean and related interactive approaches further show that human guidance remains important when cleaning choices affect downstream analysis \citep{krishnan2016activeclean}.

Knowledge-based and retrieval-based approaches are especially relevant when table content is insufficient to interpret a value. KATARA uses knowledge bases and crowdsourcing to support cleaning decisions \citep{chu2015katara}; RetClean uses retrieval to support LLM-assisted cleaning \citep{naeem2024retclean}; and retrieval-augmented generation separates retrieved evidence from generated output \citep{lewis2020rag}. Work on attribution and fact verification motivates explicit checks that a cited source supports the generated claim \citep{rashkin2023measuring,thorne2018fever}. Provenance research further motivates recording why and where a decision originated, including data, rules, tools, and evidence \citep{buneman2001provenance,cheney2009provenance,provdm2013}.

Recent LLM and agent research extends these ideas with reasoning and tool use. ReAct and Toolformer show how language models can combine reasoning with external actions or tools \citep{yao2023react,schick2023toolformer}. In data preparation, foundation models and LLM-based systems can wrangle tables, profile semantics, generate workflows, and standardise values \citep{narayan2022foundation,li2025autodcworkflow,qi2024cleanagent,huang2024cocoon}. However, empirical studies also show sensitivity to prompts, domain context, and verification mechanisms \citep{biester2024llmclean,zhang2024datacleaningllm}. The gap addressed here is therefore not whether LLMs can propose cleaning actions, but how profiling, executable tools, evidence controls, and conservative repair policies affect cleaning decisions when no clean reference is available.

Two implications follow from this literature. First, reference-free cleaning requires an explicit distinction between detecting a possible issue and modifying a value. Detection can be supported by profiling, constraints, and anomaly indicators, but repair requires stronger evidence because a value may be unusual and still valid. Second, evidence grounding must be operationalised rather than asserted. A system that retrieves a source or produces a citation is not necessarily source-grounded unless the final decision can be traced to evidence that is relevant, specific, current, and consistent with the proposed rule. These implications motivate the taxonomies, framework, and experimental design used in the present study.

\section{Research Questions and Hypotheses}
\label{sec:research_questions}

The study addresses five research questions: \textbf{RQ1} asks how different baseline and agent configurations affect reference-free data-cleaning behavior; \textbf{RQ2} asks which tool capabilities change cleaning decisions; \textbf{RQ3} asks whether source-ranking and citation-alignment controls reduce unsupported rules or unsafe repairs; \textbf{RQ4} asks whether a conservative repair policy improves valid-signal preservation and reduces unnecessary modification; and \textbf{RQ5} asks what operational cost is introduced as additional agent capabilities are enabled.

The corresponding hypotheses are that profiling and executable validation improve structural and syntactic issue detection (H1); source-ranking and citation-alignment controls reduce unsupported-rule behavior but may not guarantee strong citation support for every decision (H2); conservative repair policies reduce unsafe repairs and unnecessary modifications (H3); the full conservative evidence-grounded configuration shifts behavior toward more conservative and auditable decisions but may introduce trade-offs in detection, direct repair, runtime, and token cost (H4); and the relative effectiveness of agent configurations depends on dataset characteristics and cleaning requirements (H5).

\section{Problem Definition}
\label{sec:problem}

Let $X$ denote the raw dataset. It may be accompanied by schema $S$, optional metadata $M$, sample records $R$, profiling summaries $P$, and optional documentation $G$. Following context-driven data-quality assessment, the dataset is associated with a structured cleaning context
\begin{equation}
\Gamma_X=\langle C_{src}, C_{str}, C_{sem}, C_{temp}, C_{use}, C_{risk}, C_{policy}, C_{prov}\rangle,
\end{equation}
where the components represent source, structural, semantic, temporal, use, risk, policy, and provenance characteristics. In supervised evaluation, a clean reference table $X^{*}$ may be available. In the reference-free setting considered here, however, $X^{*}$ is unavailable. The agent must therefore infer cleaning requirements from the available decision context
\begin{equation}
K = \{X, S, M, R, P, G, \Gamma_X, E_{ext}, T\},
\end{equation}
where $E_{ext}$ denotes retrieved evidence and $T$ denotes tool outputs such as parsing, duplicate, unit, temporal, and consistency checks. The set $K$ represents the information and outputs that may support a cleaning decision; it does not imply that every component is required in every cleaning task. In the completed experiment, retrieval used a controlled local evidence corpus rather than online retrieval.

The agent produces cleaning decisions
\begin{equation}
\mathcal{C} = \{c_1, c_2, \ldots, c_n\},
\end{equation}
with each decision represented as
\begin{equation}
c_i = \langle \text{issue}_i, \text{evidence}_i, \text{requirement}_i, \text{rule}_i, \text{action}_i, \text{confidence}_i, \text{provenance}_i \rangle.
\end{equation}
The action belongs to
\begin{equation}
\begin{array}{l}
\mathcal{A}=\{\text{safe-repair}, \text{conditional-repair}, \text{flag-only}, \\
\quad \text{preserve}, \text{human-review}, \text{reject-repair}\}.
\end{array}
\end{equation}
This notation separates the raw dataset $X$ from the decision set $\mathcal{C}$. A repair is permitted only when the evidence is sufficient, the rule is operationally checkable, the transformation is reversible, and the conservative cleaning policy is not violated. The formulation also separates a cleaning requirement from an applied transformation. A requirement may state that a timestamp should be parseable, a code should belong to a documented set, or a measurement should respect a unit convention, but the corresponding action may still be flagging or preservation when the evidence is incomplete. This separation is central to avoiding unsupported repair in the absence of $X^{*}$.

The framework follows six principles: repairs should be evidence-grounded; rare or extreme values should not be modified solely because they are unusual; executable checks should be used where deterministic validation is more reliable than natural-language judgement; external sources should be ranked and aligned with the decisions they support; transformations should preserve provenance and reversibility; and the effects of added agent capabilities should be evaluated empirically rather than assumed beneficial.

\section{Knowledge and Decision Taxonomies}
\label{sec:taxonomies}

The proposed framework uses five taxonomies to specify what the agent should produce, what contextual knowledge may support its decisions, which tools it can use, which evidence can support decisions, and which conservative action classes are available. These taxonomies synthesize prior work on multidimensional data quality, context-aware assessment, profiling, knowledge-based cleaning, retrieval, attribution, provenance, constraint-based repair, probabilistic cleaning, and anomaly detection. Data-quality and context-aware assessment work motivate the schema, semantic, usage, policy, and risk categories \citep{wang1996beyond,batini2006dataquality}. Profiling and classical cleaning systems motivate tool-mediated detection outputs and executable checks \citep{rahm2000cleaning,abedjan2016detecting}. Knowledge-based and retrieval-based systems motivate the evidence taxonomy \citep{chu2015katara,lewis2020rag}, while attribution and provenance research motivate citation alignment and decision traceability \citep{rashkin2023measuring,buneman2001provenance}. Constraint-based repair, probabilistic cleaning, and anomaly-detection research motivate the conservative distinction between repair, flagging, preservation, review, and rejection \citep{bohannon2005cost,rekatsinas2017holoclean,chandola2009anomaly}. The aim is not to claim that every category is necessary or equally important for every dataset, but to make the knowledge, capabilities, evidence, and decision requirements of the framework explicit.

\subsection{Taxonomy-to-Implementation Mapping}
\label{subsec:taxonomy_implementation}

Table~\ref{tab:taxonomy_implementation_map} summarizes how the taxonomy is operationalized in the reported ablation. This mapping is included to avoid overclaiming: some categories are directly measured, some are partly implemented through context and logging, and others define framework-level concepts or referral mechanisms rather than independently adjudicated experimental variables.

\begingroup
\footnotesize
\setlength{\tabcolsep}{3pt}
\renewcommand{\arraystretch}{1.0}
\begin{longtable}{P{0.18\textwidth}P{0.27\textwidth}P{0.23\textwidth}P{0.24\textwidth}}
\caption{Mapping between taxonomy components and the reported experimental implementation.}
\label{tab:taxonomy_implementation_map}\\
\toprule
\textbf{Taxonomy component} & \textbf{Framework role} & \textbf{Implementation status} & \textbf{Configurations and evaluation}\\
\midrule
\endfirsthead
\toprule
\textbf{Taxonomy component} & \textbf{Framework role} & \textbf{Implementation status} & \textbf{Configurations and evaluation}\\
\midrule
\endhead
Required outputs & Expose issues, requirements, rules, decisions, evidence, provenance, scripts, and audit summaries. & Logged in the output schema; capability-specific fields are absent when not enabled. & A0--A6; checked through counts, repair metrics, schema/JSON validity, reproducibility, and auditability.\\
Prior knowledge & Provide schema, unit, temporal, policy, purpose, and risk context that may support repair decisions. & Partly implemented through context profiles, schema/samples, profiling summaries, and risk descriptions; not separately ablated by knowledge type. & Context is available to different degrees across configurations, but individual knowledge types are not evaluated independently.\\
Tool capabilities & Delegate profiling, parsing, validation, temporal/constraint checks, and reversible scripts to tools. & Implemented for profiling/checking in A1 and for executable validation/script generation from A3 onward. & A1 tests deterministic profiling; A3--A6 test executable checks and reversible scripts; evaluated through detection, repair, execution, and operational metrics.\\
Evidence and citation controls & Distinguish evidence types, rank sources, and align cited claims with decisions. & Controlled local evidence retrieval is used in A4; source-ranking and citation-alignment checks are used in A5--A6. & A4 tests evidence availability; A5--A6 test ranking/alignment; evaluated through evidence coverage, unsupported-rule rate, citation alignment, runtime, and token use.\\
Decision classes and conservative policy & Convert suspected issues into repair, flagging, preservation, human review, or rejected repair. & Decision labels are logged across configurations; explicit conservative policy and provenance control are enabled most fully in A6. & A5--A6 isolate conservative control; evaluated through repair/safety metrics, valid-signal preservation, human-review rate, and reproducibility.\\
Human review and expert adjudication & Provide a high-authority pathway for ambiguous or high-risk cases. & Implemented only as a referral/decision class; no external expert adjudication of original-data cases was performed. & Mainly visible in A6 original-data behavior; reported descriptively through referral rate and discussed as a limitation/future-work need.\\
\bottomrule
\end{longtable}
\endgroup

\subsection{Required-Output Taxonomy}
\label{subsec:required_outputs}

The required-output taxonomy in Table~\ref{tab:required_outputs} defines the artefacts that must be produced for an auditable cleaning decision. It separates candidate issues, requirements, rules, decisions, repairs, provenance, and escalation outputs so that a detected anomaly is not automatically treated as a repairable error.

\begin{longtable}{P{0.23\textwidth}P{0.48\textwidth}P{0.21\textwidth}}
\caption{Required outputs of the evidence-grounded cleaning agent.}
\label{tab:required_outputs}\\
\toprule
\textbf{Output} & \textbf{Definition} & \textbf{Purpose}\\
\midrule
\endfirsthead
\toprule
\textbf{Output} & \textbf{Definition} & \textbf{Purpose}\\
\midrule
\endhead
Issue inventory & Structured list of suspected quality issues, including affected columns, rows, entities, timestamps, and issue contexts. & Identifies what may be wrong.\\
Issue type & Classification of the issue, such as missing value, duplicate, invalid unit, impossible value, inconsistent category, temporal gap, entity mismatch, or suspicious anomaly. & Supports systematic analysis.\\
Evidence record & Dataset-derived or externally retrieved evidence used to support the issue or repair decision. & Reduces hallucinated cleaning.\\
Cleaning requirement & A requirement inferred from schema, profiling, documentation, standards, or authoritative sources. & Defines what the data should satisfy.\\
Validation rule & A machine-checkable rule derived from a cleaning requirement. & Enables repeatable validation.\\
Repair proposal & A suggested transformation, imputation, normalization, deletion, or correction. & Generates candidate repairs when safe.\\
Decision class & Final action category: safe repair, conditional repair, flag only, preserve, human review, or reject repair. & Prevents over-cleaning.\\
Confidence and evidence strength & Numeric or ordinal estimate of decision reliability based on evidence quality and rule support. & Supports prioritization.\\
Provenance log & Record of prompt context, source, tool output, rule, affected records, and transformation. & Ensures auditability.\\
Reversible script & Script that applies approved transformations while preserving original values. & Enables rollback and review.\\
Audit report & Human-readable summary of issues, decisions, repairs, rejected repairs, evidence quality, and remaining uncertainties. & Supports scientific reporting.\\
\bottomrule
\end{longtable}

These outputs support auditability. A reviewer, data owner, or downstream analyst should be able to see what issue was detected, what requirement was inferred, which evidence supported it, what action was chosen, and whether any value was changed. The taxonomy therefore treats provenance and escalation as first-class outputs rather than optional metadata.

\subsection{Prior-Knowledge Taxonomy}
\label{subsec:prior_knowledge}

The prior-knowledge taxonomy in Table~\ref{tab:prior_knowledge} identifies contextual information that may support the interpretation of a dataset before repair decisions are made. It includes schema, semantic, temporal, entity, risk, policy, and provenance knowledge because reference-free decisions may depend on both data-derived patterns and external interpretation. The current experiment does not evaluate these knowledge categories independently.

\begin{longtable}{P{0.24\textwidth}P{0.43\textwidth}P{0.25\textwidth}}
\caption{Taxonomy of prior knowledge required by an autonomous cleaning agent.}
\label{tab:prior_knowledge}\\
\toprule
\textbf{Knowledge Type} & \textbf{Definition} & \textbf{Domain-Independent Example}\\
\midrule
\endfirsthead
\toprule
\textbf{Knowledge Type} & \textbf{Definition} & \textbf{Domain-Independent Example}\\
\midrule
\endhead
Schema knowledge & Column names, data types, key fields, expected structural constraints, and valid formats. & Identifier fields, timestamps, categorical codes, numeric measures.\\
Unit and scale knowledge & Meaning, unit, scale, and compatibility of measured values. & Currency units, medical units, sensor units, percentages, rates.\\
Temporal knowledge & Expected frequency, ordering, allowable time gaps, duplicate timestamps, and continuity constraints. & Hourly readings, admission/discharge dates, transaction sequences.\\
Entity knowledge & Entities represented in the dataset and valid relationships among them. & Patient ID, station ID, customer ID, product ID, location ID.\\
Domain-threshold knowledge & Expected ranges, warning limits, legal limits, operational thresholds, or plausible bounds. & Valid age range, sensor range, maximum transaction amount.\\
Contextual-event knowledge & External events or domain conditions that may explain unusual values. & Maintenance event, market shock, public-health event, environmental event.\\
Missingness knowledge & Meaning of nulls, placeholders, censored values, unavailable observations, and not-applicable values. & ``NA'', ``unknown'', below-detection values, empty codes.\\
Constraint knowledge & Functional, relational, temporal, arithmetic, or semantic constraints that should hold. & End date follows start date; total equals sum of parts.\\
Cleaning-policy knowledge & Rules defining when to repair, flag, preserve, or escalate. & Do not modify high-impact values without evidence.\\
Evidence-provenance knowledge & Authority, date, source type, citation alignment, and source reliability. & Official documentation preferred over unsupported web pages.\\
Task-purpose knowledge & Intended downstream use of the data, including whether cleaning is for reporting, modeling, compliance, or exploration. & Preserve anomalies for anomaly detection; normalize for reporting.\\
Risk knowledge & Consequences of false repair, false preservation, deletion, or imputation. & Clinical, financial, safety, regulatory, or scientific impact.\\
\bottomrule
\end{longtable}

This taxonomy is deliberately broader than schema metadata because many cleaning decisions may depend on semantic, temporal, or policy context. For example, a radiation spike, clinical measurement, or financial transaction cannot always be judged safely from column names alone. Interpreting such values may require information about how the data were produced, how they are used, and what risks are associated with an incorrect repair or a missed issue.

\subsection{Tool-Capability Taxonomy}
\label{subsec:tools}

The tool-capability taxonomy in Table~\ref{tab:tools} distinguishes computations that should be delegated to deterministic or executable tools from judgements that require evidence-aware reasoning. Tool outputs are treated as evidence records rather than as unverified suggestions.

\begin{longtable}{P{0.23\textwidth}P{0.45\textwidth}P{0.24\textwidth}}
\caption{Tool-capability taxonomy for evidence-grounded data cleaning.}
\label{tab:tools}\\
\toprule
\textbf{Tool Class} & \textbf{Function} & \textbf{Example Output}\\
\midrule
\endfirsthead
\toprule
\textbf{Tool Class} & \textbf{Function} & \textbf{Example Output}\\
\midrule
\endhead
Profiling tools & Summarize types, missingness, uniqueness, distributions, outliers, patterns, and correlations. & Missingness rate, duplicate count, value ranges.\\
Parsing and type tools & Validate dates, numbers, codes, units, regular expressions, and encodings. & Invalid timestamp list.\\
Duplicate and entity-resolution tools & Detect exact duplicates, near duplicates, and possible entity conflicts. & Candidate duplicate clusters.\\
Temporal validation tools & Check ordering, gaps, repeated timestamps, rolling statistics, and frequency expectations. & Gap intervals and repeated measurements.\\
Constraint-checking tools & Execute functional, arithmetic, relational, and semantic constraints. & Rule violation table.\\
Unit and conversion tools & Validate unit compatibility and perform documented conversions. & Converted values with original unit preserved.\\
Controlled evidence-retrieval tools & Retrieve supporting evidence from a controlled evidence collection; online evidence retrieval can be used in future replications when source volatility and reproducibility are explicitly controlled. & Candidate evidence sources.\\
Source-ranking tools & Score retrieved sources by authority, relevance, recency, specificity, and consistency. & Ranked evidence list.\\
Citation-alignment tools & Check whether a cited statement supports a generated requirement or repair. & Supported, partially supported, unsupported.\\
Script-generation tools & Generate reversible cleaning scripts and execution logs. & Python or SQL transformation script.\\
Script-execution tools & Run generated validation or repair scripts in a controlled environment and capture errors, modified records, rollback information, and execution status. & Execution log, error trace, affected-row count, rollback file.\\
\bottomrule
\end{longtable}

The taxonomy assigns tools to verification roles rather than treating the LLM as the sole decision-maker. Deterministic checks are appropriate for parsing, type validation, duplicate detection, and script execution, while the agent is responsible for interpreting those outputs with respect to domain context and repair risk.

\subsection{Evidence Taxonomy}
\label{subsec:evidence_taxonomy}

For evidence item $e$, the framework represents evidence strength as
\begin{equation}
ES(e)=\alpha A(e)+\beta R(e)+\gamma Sp(e)+\delta C(e)+\eta T(e),
\end{equation}
where $A(e)$ denotes source authority, $R(e)$ relevance to the cleaning decision, $Sp(e)$ specificity, $C(e)$ consistency with other available evidence, and $T(e)$ temporal validity. The coefficients $\alpha$, $\beta$, $\gamma$, $\delta$, and $\eta$ represent the relative importance assigned to these dimensions. The formulation provides a structured way to compare evidence sources; it is not intended as a universal evidence-quality model.

\begin{longtable}{P{0.24\textwidth}P{0.45\textwidth}P{0.23\textwidth}}
\caption{Evidence taxonomy for cleaning decisions.}
\label{tab:evidence_taxonomy}\\
\toprule
\textbf{Evidence Type} & \textbf{Definition} & \textbf{Typical Strength}\\
\midrule
\endfirsthead
\toprule
\textbf{Evidence Type} & \textbf{Definition} & \textbf{Typical Strength}\\
\midrule
\endhead
Schema evidence & Evidence from names, declared types, primary keys, foreign keys, and constraints. & Medium to high.\\
Profile evidence & Evidence from distributions, missingness, duplicates, type errors, and statistical summaries. & Medium; weaker for semantic repair.\\
Executable evidence & Evidence from deterministic validation, parsing, rule execution, and reproducible checks. & High for syntactic and structural issues.\\
Dataset documentation & Official metadata, data dictionary, README, or provider documentation. & High when specific and current.\\
Institutional or regulatory source & Government, standards body, organizational documentation, or domain authority. & High when directly relevant.\\
Scientific literature & Peer-reviewed source providing domain thresholds, measurement interpretation, or methods. & Medium to high depending on specificity.\\
Technical manual & Device, software, or system manual explaining units, codes, or valid ranges. & High for instrumentation-specific rules.\\
General web source & Blog, forum, or unverified page. & Low; should not justify high-impact repairs alone.\\
Human expert review & Domain expert judgment or adjudication. & High, but may be subjective and should be logged.\\
\bottomrule
\end{longtable}

The evidence-strength equation is not intended as a universal scoring formula. It defines the dimensions that should be logged and made explicit when a decision depends on evidence. Different domains may weight the dimensions differently; for example, clinical and regulatory settings may require stronger authority and specificity than exploratory data analysis.

\subsection{Cleaning Decision Taxonomy}
\label{subsec:decision_classes}

The decision classes in Table~\ref{tab:decision_classes} are defined by evidence sufficiency and repair risk. This taxonomy is the main bridge between the conceptual problem and the executable experiment: it determines whether a candidate issue becomes a repair, a flag, a preservation decision, a human-review referral, or a rejected repair proposal. Because no reference value is available, the taxonomy gives the agent conservative alternatives to modification: flagging, preserving possible valid signals, escalating to human review, or rejecting unsupported repair.

\begin{longtable}{P{0.18\textwidth}P{0.36\textwidth}P{0.38\textwidth}}
\caption{Cleaning decision classes used by the proposed framework.}
\label{tab:decision_classes}\\
\toprule
\textbf{Decision Class} & \textbf{Definition} & \textbf{Example}\\
\midrule
\endfirsthead
\toprule
\textbf{Decision Class} & \textbf{Definition} & \textbf{Example}\\
\midrule
\endhead
Safe repair & The issue is clearly erroneous and the correction is directly supported by rules, documentation, executable checks, or strong evidence. & Standardizing timestamp format; replacing documented placeholder codes with null.\\
Conditional repair & A repair is plausible but depends on an explicit documented assumption. & Converting units when the source unit is strongly inferred but not directly stated.\\
Flag only & The value is suspicious but evidence is insufficient for modification. & Sudden spike, unusual transaction, or unexpected sensor value.\\
Preserve as possible valid signal & The value is unusual but may represent a meaningful rare event or domain signal. & Rare clinical result, extreme environmental reading, valid financial shock.\\
Needs human review & The issue is high-risk, ambiguous, or unresolved by available evidence. & Conflicting evidence or uncertain domain interpretation.\\
Reject unsupported repair & A repair proposal lacks sufficient support or violates policy. & Replacing an outlier only because it is statistically extreme.\\
\bottomrule
\end{longtable}

\section{Methodology: Evidence-Grounded Agentic Cleaning Framework}
\label{sec:framework}

\begin{figure}[H]
    \centering
    \includegraphics[width=\textwidth]{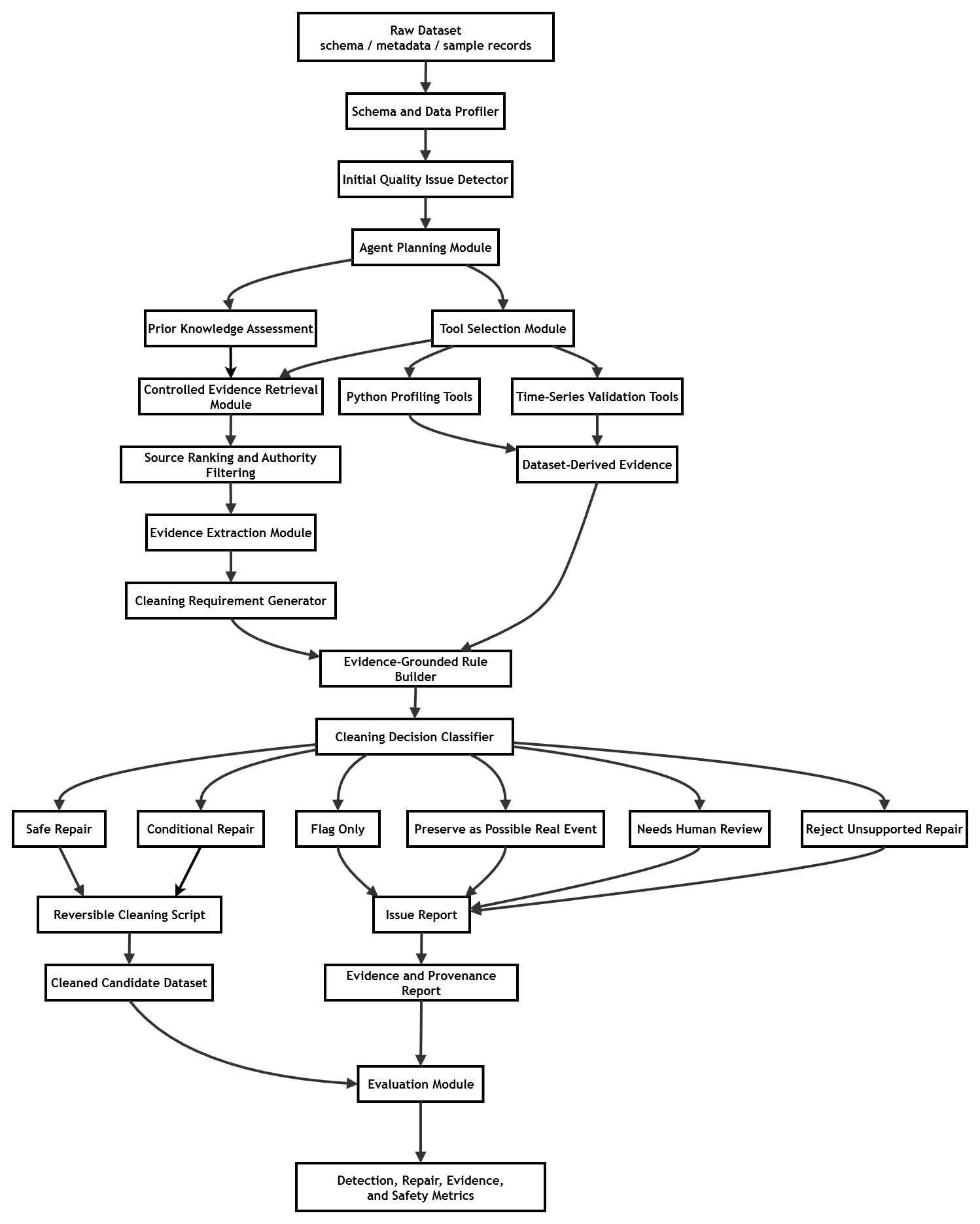}
    \caption{Architecture of the evidence-grounded agentic data-cleaning framework.}
    \label{fig:cleaning_flow}
\end{figure}

The framework translates the problem definition into a multi-stage process: issue detection, evidence acquisition, requirement generation, rule construction, decision classification, reversible repair execution, and provenance logging. The process begins with the raw dataset, schema, metadata, and sample records. Profiling derives structural and statistical summaries; the issue detector identifies candidate quality problems; and the agent planner decides whether internal evidence is sufficient or whether additional tool use or controlled evidence retrieval is required. Figure~\ref{fig:cleaning_flow} summarises the architecture.

The framework distinguishes dataset-derived evidence from retrieved evidence. Dataset-derived evidence includes schema constraints, profiling summaries, missingness, duplicates, temporal behavior, distributions, and executable validation results. Retrieved evidence includes documentation, standards, manuals, domain references, or other configured sources. Both forms of evidence can support the requirements and rules used to evaluate candidate cleaning decisions.

The agent workflow has five steps. First, it constructs $\Gamma_X$ and profiles the dataset. Second, it proposes candidate issues and identifies the contextual knowledge relevant to their interpretation. Third, it invokes tools for deterministic checks where applicable. Fourth, it creates evidence-backed requirements and rules, with evidence assessed according to authority, relevance, specificity, consistency, and temporal validity. Fifth, it classifies each decision into the conservative action set and generates reversible scripts only when the policy allows repair. All decisions record provenance, including evidence identifiers, rule text, tool outputs, and the original value.

The conservative repair policy acts as a control layer over candidate transformations. A decision may be classified as safe repair only when the candidate issue is clear, the rule is supported by available evidence, and the transformation can be reversed. When evidence is partial, conflicting, or high risk, the policy favours conditional repair, flagging, preservation, human review, or rejection of unsupported repair. This approach may reduce direct repair because uncertain cases can be preserved or escalated instead of being modified automatically.

The framework also separates planning from execution. The agent may propose a requirement or rule, but executable checks and reversible scripts determine whether the proposal can be operationalised. This reduces reliance on fluent natural-language explanations when deterministic validation is available. In the same way, evidence records are not treated as decorations for the report; they are intended to be linked to requirements and actions so that reviewers can distinguish supported repairs from unsupported or ambiguous suggestions.

When a repair is executed, the framework uses reversible scripts that preserve original values and produce logs that can be inspected after execution. This separates the agent's recommendation from the deterministic effect of the generated transformation and supports auditability and reproducibility.

\section{Experimental Design and Implementation}
\label{sec:experimental_design}

The exploratory experiment evaluates seven configurations across three datasets and two run modes. A0 and A1 serve as comparison baselines, while A2--A6 form the progressive LLM-based ablation. Three repeated runs were performed for each dataset--mode--configuration combination, resulting in 126 completed runs. The datasets represent financial decision support \citep{german_credit}, clinical records \citep{ehr_dataset}, and environmental radiation monitoring \citep{fadlallah2018raden}. Each dataset was supplied with a context profile describing its domain, operational setting, schema, risk level, and relevant cleaning context. Table~\ref{tab:datasets_context} summarises the dataset roles.

\begin{table}[htbp]
\centering
\caption{Datasets and context profiles used in the experimental study.}
\label{tab:datasets_context}
\footnotesize
\resizebox{\textwidth}{!}{%
\begin{tabular}{P{0.15\textwidth}P{0.15\textwidth}P{0.20\textwidth}P{0.5\textwidth}}
\toprule
\textbf{Dataset} & \textbf{Domain} & \textbf{Cleaning Context} & \textbf{Main Context Requirements}\\
\midrule
German Credit & Finance & Financial decision support and credit-risk interpretation. & Schema and categorical-code interpretation, plausible financial ranges, sensitive attributes, target-label protection, and conservative handling of unusual but valid credit profiles.\\
EHR & Healthcare & Clinical data-quality assessment. & Patient/entity consistency, clinical attribute validation, missingness interpretation, categorical consistency, high-risk repair control, and human-review escalation.\\
Radiation & Environmental monitoring / IoT & Sensor monitoring and anomaly detection. & Timestamp parsing, temporal gaps, duplicate measurements, unit and scale interpretation, sensor anomalies, battery/status fields, and preservation of plausible radiation spikes.\\
\bottomrule
\end{tabular}%
}
\end{table}

The controlled synthetic-corruption mode injects known errors and plausible valid-signal cases, allowing detection, repair, and preservation metrics to be computed against ground truth. The original-data mode applies the same configurations to the unmodified datasets. Because the original datasets do not have a trusted clean reference or manual adjudication in the verified outputs, original-data results are descriptive only. Table~\ref{tab:synthetic_corruption_protocol} summarises the corruption and valid-signal protocol.

\begin{table}[H]
\centering
\caption{Synthetic corruption and valid-signal protocol used in the completed experiment. Counts are per dataset and repeated run; the same ground-truth file for a given dataset and run was used across all A0--A6 configurations.}
\label{tab:synthetic_corruption_protocol}
\scriptsize
\resizebox{\textwidth}{!}{%
\begin{tabular}{lrrrrrrrrrrr}
\toprule
\textbf{Dataset} &
\textbf{Missing} &
\textbf{Type} &
\textbf{Range} &
\textbf{Invalid} &
\textbf{Invalid} &
\textbf{Duplicate} &
\textbf{Entity} &
\textbf{Missing} &
\textbf{Injected} &
\textbf{Valid} &
\textbf{Total} \\
&
\textbf{value} &
\textbf{error} &
\textbf{violation} &
\textbf{category} &
\textbf{timestamp} &
\textbf{row} &
\textbf{duplicate} &
\textbf{identifier} &
\textbf{errors} &
\textbf{signals} &
\textbf{records} \\
\midrule
German Credit & 10 & 5 & 5 & 5 & 0 & 5 & 0 & 0 & 30 & 5 & 35 \\
EHR & 10 & 5 & 5 & 5 & 0 & 5 & 2 & 1 & 33 & 5 & 38 \\
Radiation & 10 & 5 & 5 & 0 & 5 & 5 & 0 & 0 & 30 & 5 & 35 \\
\bottomrule
\end{tabular}%
}
\end{table}

The experiment uses two comparison baselines followed by a progressive capability ablation. A0 is a schema-and-sample LLM baseline without profiling, executable tools, evidence retrieval, or conservative repair controls. A1 is a deterministic profiling-and-checking baseline and does not use an LLM. A2 combines profiling summaries with LLM reasoning and serves as the starting point of the progressive LLM-based sequence. A3 adds executable Python checks and reversible-script generation, A4 adds controlled evidence retrieval, A5 adds source ranking and citation-alignment checks, and A6 adds the conservative repair policy.

A0 and A1 should therefore be interpreted as contrasting baselines rather than consecutive steps of a single-factor ablation. The more direct capability comparisons are A2 to A3 for executable tools, A3 to A4 for evidence retrieval, A4 to A5 for source-ranking and citation-alignment controls, and A5 to A6 for the conservative repair policy.

The LLM-based configurations used OpenRouter with \texttt{openai/gpt-oss-20b}, while A1 was deterministic and did not use an LLM. The temperature was fixed at 0.2 and the maximum output length at 3000 tokens. The prompt context included the dataset schema, metadata, context profile, up to 10 sample records, and the profiling, tool, or evidence information enabled for each configuration. Model settings, retry logic, output constraints, and postprocessing were kept fixed across all LLM-based configurations.

The output followed a fixed JSON schema containing candidate issues, requirements, validation rules, decisions, evidence references, provenance information, and audit indicators. This common schema made outputs comparable across configurations and allowed systematic postprocessing without manually interpreting free-form responses. When a response was malformed, the saved postprocessing pipeline applied schema normalization or JSON repair where needed, and these events were retained in the execution audit. The experiment was not rerun after analysis, and all reported results were computed from the saved postprocessed artifacts.

\begin{table}[H]
\centering
\caption{Controlled local evidence corpus used in the completed experiment.}
\label{tab:local_evidence_corpus}
\footnotesize
\resizebox{\textwidth}{!}{%
\begin{tabular}{l r p{0.58\textwidth}}
\toprule
\textbf{Corpus file} & \textbf{Records} & \textbf{Main contents} \\
\midrule
\texttt{common.jsonl} & 2 & Reference-free cleaning policy and provenance/reversibility policy shared across datasets. \\
\texttt{german\_credit.jsonl} & 5 & German Credit dataset card, UCI documentation metadata, financial-domain cleaning policy, valid ranges, and valid-signal preservation policy. \\
\texttt{ehr\_dataset.jsonl} & 5 & EHR dataset card, Kaggle documentation metadata, clinical safety policy, identifier policy, and categorical-value policy. \\
\texttt{radiation\_dataset.jsonl} & 5 & Radiation dataset card, radiation-monitoring context, anomaly-preservation policy, sensor validity checks, and temporal-consistency policy. \\
\midrule
Total & 17 & Fixed project-local evidence records used by A4--A6 in the reported experiment. \\
\bottomrule
\end{tabular}%
}
\end{table}

For A5 and A6, retrieved evidence was ranked using the evidence-strength formulation introduced in Section~\ref{sec:taxonomies}. The implemented weights were 0.30 for authority, 0.25 for relevance, 0.20 for specificity, 0.15 for consistency, and 0.10 for temporal validity. These weights were fixed for all source-ranked runs and were used as an operational ranking heuristic rather than as a universally validated evidence-quality model. In the controlled local corpus, consistency was not separately annotated and therefore used a fixed value of 0.7 for all evidence records. Consequently, the consistency component did not differentiate evidence sources in the reported experiment.

\section{Evaluation Protocol}
\label{sec:evaluation}

The evaluation separates detection and repair effectiveness, safety, evidence grounding, operational cost, and reproducibility. Synthetic-corruption runs support detection, repair, and valid-signal preservation metrics because injected ground truth is available. Original-data runs are interpreted descriptively because no trusted clean reference or manual adjudication is available. Execution indicators such as retries, JSON repair, schema normalization, and reversible-script execution are reported separately as an audit and are not treated as measures of cleaning effectiveness.

\begin{longtable}{P{0.15\textwidth}P{0.25\textwidth}P{0.5\textwidth}}
\caption{Evaluation metrics for reference-free evidence-grounded cleaning.}
\label{tab:metrics}\\
\toprule
\textbf{Metric Group} & \textbf{Metric} & \textbf{Definition}\\
\midrule
\endfirsthead
\toprule
\textbf{Metric Group} & \textbf{Metric} & \textbf{Definition}\\
\midrule
\endhead
Detection & Precision & In the controlled synthetic-corruption setting, the fraction of detected issues that match injected error ground truth. Original-data precision is not interpreted because no injected or manually validated ground truth is available.\\
Detection & Recall & In the controlled synthetic-corruption setting, the fraction of injected errors detected by the agent. Original-data recall is not interpreted because no ground-truth error set is available.\\
Detection & F1-score & Harmonic mean of synthetic detection precision and recall.\\
Repair & Expected-repair match rate & Among repair decisions matched to injected errors for which repair is expected, the fraction whose proposed repair contains the known expected correction.\\
Repair & Safe repair rate & Fraction of repair-class decisions that match synthetic ground-truth cases for which repair is appropriate, excluding repairs applied to valid-signal or non-repair cases and repairs with no matching ground-truth issue.\\
Repair & Unsafe repair rate & Fraction of repair-class decisions applied to synthetic cases marked as valid signals or cases whose expected action is preservation, flagging, or human review.\\
Repair & Unnecessary modification rate & Fraction of repair-class decisions whose associated issue does not match any synthetic ground-truth case.\\
Evidence & Evidence coverage & Fraction of decisions that reference at least one evidence record available in the agent output.\\
Evidence & Mean evidence-strength score & Mean evidence-strength value across the evidence records available in the agent output, including profiling, tool, and controlled retrieved evidence.\\
Evidence & Unsupported-rule rate & Fraction of generated validation rules that do not reference available evidence.\\
Evidence & Citation alignment & Mean, across decisions, of the maximum Jaccard similarity between the decision text and its cited evidence statements, reported for configurations with citation-alignment checks enabled.\\
Safety & Conservative flag rate & Fraction of decisions classified as flag-only, preserve, human-review, or reject-repair.\\
Safety & Valid-signal preservation rate & Among injected valid-signal cases detected as issues, the fraction assigned to preserve, flag-only, or human-review rather than repair.\\
Safety & Human-review referral rate & Fraction of decisions escalated for human review.\\
Operational & Runtime & Per-run execution time recorded in seconds and aggregated by dataset, mode, and configuration.\\
Operational & Token use & Prompt, completion, and total token usage recorded for LLM-based runs.\\
Operational & Reproducibility & Pairwise exact agreement across repeated runs of decision signatures constructed from issue identifiers and their assigned decision classes within each dataset--mode--configuration group.\\
\bottomrule
\end{longtable}

Table~\ref{tab:baselines} summarises the roles and limitations of A0--A6. A0 and A1 provide contrasting LLM and deterministic baselines and are not treated as a single-factor ablation. A2 provides the profiling-aware LLM starting point for the progressive capability analysis. The more direct comparisons are A2 versus A3 for executable Python validation, A3 versus A4 for controlled evidence retrieval, A4 versus A5 for source ranking and citation alignment, and A5 versus A6 for the conservative repair policy.

\begingroup
\footnotesize
\setlength{\tabcolsep}{3pt}
\renewcommand{\arraystretch}{1.02}
\begin{longtable}{P{0.22\textwidth}P{0.47\textwidth}P{0.23\textwidth}}
\caption{Comparison baselines and progressive ablation configurations.}
\label{tab:baselines}\\
\toprule
\textbf{Configuration} & \textbf{Implemented description} & \textbf{Expected limitation}\\
\midrule
\endfirsthead
\toprule
\textbf{Configuration} & \textbf{Implemented description} & \textbf{Expected limitation}\\
\midrule
\endhead
A0: Schema-only LLM baseline &
LLM with schema and sample records only; no profiling summaries, executable checks, evidence retrieval, ranking/alignment, conservative policy, reversibility, or provenance logging. &
May hallucinate requirements, miss data-derived patterns, or overinterpret values without profiling and executable evidence.\\

A1: Deterministic profiling-and-checking baseline &
Deterministic profiling, missingness analysis, duplicate detection, type inference, range checks, and provenance logging; no LLM call, retrieval, ranking/alignment, conservative policy, or reversible scripts. &
Can detect structural/statistical patterns efficiently, but may treat statistical anomalies as errors without semantic, contextual, or conservative reasoning.\\

A2: LLM with profiling summaries &
LLM reasoning over profiling summaries with provenance logging; no executable Python validation, retrieval, ranking/alignment, conservative policy, or reversible scripts. &
May support explanation and contextual reasoning, but still lacks executable validation and retrieved evidence support.\\

A3: LLM with profiling and Python tools &
Adds executable Python checks for validation, parsing, duplicates, consistency, reversible-script generation, and provenance logging; no retrieval, ranking/alignment, or conservative policy. &
Can validate candidate issues through executable checks, but may still decide without retrieved contextual evidence or source-ranking controls.\\

A4: Local-corpus evidence-retrieval configuration &
Adds retrieval from the controlled local evidence corpus while keeping profiling, Python tools, reversible scripts, and provenance logging; no source ranking or citation-alignment checks. &
Makes evidence records available, but retrieved evidence may not be strictly ranked, filtered, or aligned with final decisions.\\

A5: Source-ranked evidence configuration &
Adds source ranking, filtering, evidence extraction, and citation-alignment checks on top of A4; the full conservative repair policy is not enabled. &
Can reduce unsupported-rule behavior through evidence controls, but citation alignment may remain weak and repair behavior may still be insufficiently conservative.\\

A6: Full conservative evidence-grounded configuration &
Adds the conservative repair policy on top of A5 while retaining profiling, executable tools, controlled evidence retrieval, source ranking, citation alignment, reversibility, and provenance logging. &
More expensive and may flag, preserve, refer, or reject uncertain cases instead of performing direct repair; not expected to dominate all simpler configurations.\\
\bottomrule
\end{longtable}
\endgroup

No p-values, confidence intervals, or statistical significance tests are reported because they were not part of the completed analysis. The evaluation therefore focuses on observed means, execution-audit indicators, and exact reproducibility across repeated runs. The hypotheses are interpreted using the observed directional patterns rather than statistical significance. Original-data results are kept separate from the controlled synthetic evaluation because original-data issues may represent true errors, unusual but valid values, or undocumented conventions and cannot be scored as correct or incorrect without independent adjudication.

\section{Results}
\label{sec:results}

The results are organised around execution validity, synthetic-corruption performance, original-data descriptive behavior, evidence grounding, safety, operational cost, and reproducibility. They should be read as observed configuration trade-offs in the completed exploratory ablation. Synthetic-corruption results use injected ground truth and can support detection, repair, and valid-signal preservation metrics. Original-data results are descriptive because no trusted clean reference or manual adjudication exists.

The experiment produced the expected 126 successful runs, with no final failed runs. Table~\ref{tab:results_execution_audit} shows that JSON repair was used in 30 runs, postprocessing changed 31 saved outputs, and all 126 final outputs were schema-valid after normalization. These are implementation and reproducibility facts; no agent was rerun and no new cleaning decisions were introduced during postprocessing.

\begin{table}[!htbp]
\centering
\caption{Experiment completion and postprocessing audit.}
\label{tab:results_execution_audit}
\footnotesize
\begin{tabular}{l r}
\toprule
\textbf{Audit item} & \textbf{Verified value} \\
\midrule
Expected design & $3 \times 2 \times 7 \times 3 = 126$ runs \\
Rows in final run-level metrics & 126 \\
Rows in final aggregate metrics & 42 \\
Successful final runs & 126 \\
Failed final runs & 0 \\
Runs requiring retry before success & 4 \\
Runs with JSON repair used & 30 \\
Outputs changed by postprocessing & 31 \\
Schema-invalid outputs before normalization & 25 \\
Schema-invalid outputs after normalization & 0 \\
Final schema-valid outputs & 126/126 \\
Successful reversible-script executions & 72/72 \\
\bottomrule
\end{tabular}
\end{table}

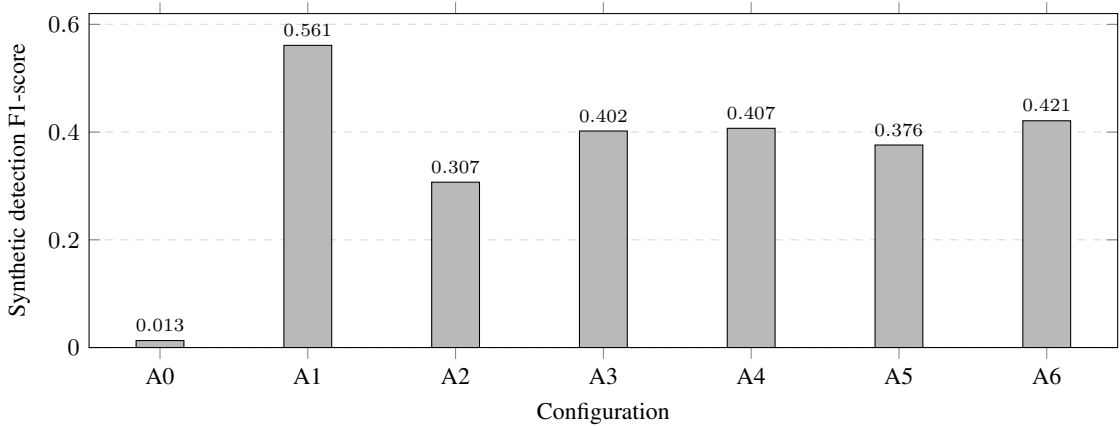
\begin{figure}[!htbp]
\centering
\begin{tikzpicture}
\begin{axis}[
    ybar,
    width=0.92\textwidth,
    height=6.0cm,
    ymin=0,
    ymax=0.62,
    ylabel={Synthetic detection F1-score},
    symbolic x coords={A0,A1,A2,A3,A4,A5,A6},
    xtick=data,
    xlabel={Configuration},
    bar width=18pt,
    enlarge x limits=0.08,
    ymajorgrids=true,
    grid style={dashed,gray!30},
    axis line style={black},
    tick label style={font=\small},
    label style={font=\small},
    nodes near coords,
    nodes near coords style={font=\scriptsize, anchor=south},
    every node near coord/.append style={
        text=black,
        /pgf/number format/fixed,
        /pgf/number format/precision=3
    },
]
\addplot+[
    fill=gray!55,
    draw=black
] coordinates {
    (A0,0.013)
    (A1,0.561)
    (A2,0.307)
    (A3,0.402)
    (A4,0.407)
    (A5,0.376)
    (A6,0.421)
};
\end{axis}
\end{tikzpicture}
\caption{Synthetic detection F1-score across the seven evaluated configurations. A0 and A1 are comparison baselines, while A2--A6 represent the progressive LLM-based capability sequence.}
\label{fig:synthetic_f1_by_config}
\end{figure}

Tables~\ref{tab:results_synthetic_detection_repair} and~\ref{tab:results_synthetic_safety} report the controlled synthetic-corruption results, while Figure~\ref{fig:synthetic_f1_by_config} provides a direct comparison of detection F1-score across configurations. A1, the deterministic profiling-and-checking baseline, achieved the highest detection F1-score at 0.561, with precision 0.687 and recall 0.484. Among the LLM-based configurations, A6 achieved the highest F1-score at 0.421, while A4 achieved the highest precision at 0.928. A0 performed poorly, with an F1-score of 0.013. Within the progressive LLM-based sequence, adding executable tools increased F1-score from 0.307 in A2 to 0.402 in A3. This pattern is consistent with H1, while the later configurations show that adding more capabilities does not consistently improve detection performance.

\begin{table}[!htbp]
\centering
\caption{Controlled synthetic-corruption detection and repair metrics by configuration.}
\label{tab:results_synthetic_detection_repair}
\scriptsize
\setlength{\tabcolsep}{4pt}
\resizebox{\textwidth}{!}{%
\begin{tabular}{l l r r r r r r r}
\toprule
\textbf{Config.} & \textbf{Configuration} & \textbf{Issues} & \textbf{Decisions} & \textbf{Prec.} & \textbf{Rec.} & \textbf{F1} & \textbf{Repair match} & \textbf{Safe repair} \\
\midrule
A0 & Schema-only & 3.222 & 3.222 & 0.074 & 0.007 & 0.013 & 0.000 & 0.000 \\
A1 & Profiling/checks & 22.556 & 22.556 & 0.687 & 0.484 & 0.561 & 0.000 & 0.770 \\
A2 & LLM + profiling & 9.444 & 6.111 & 0.727 & 0.200 & 0.307 & 0.000 & 0.000 \\
A3 & Python tools & 10.111 & 5.333 & 0.884 & 0.267 & 0.402 & 0.028 & 0.444 \\
A4 & Python + controlled evidence & 11.222 & 7.000 & 0.928 & 0.277 & 0.407 & 0.000 & 0.556 \\
A5 & Source-ranked evidence & 12.667 & 4.667 & 0.804 & 0.264 & 0.376 & 0.000 & 0.222 \\
A6 & Full conservative & 12.778 & 3.444 & 0.885 & 0.302 & 0.421 & 0.000 & 0.000 \\
\bottomrule
\end{tabular}%
}
\end{table}

\begin{table}[!htbp]
\centering
\caption{Controlled synthetic-corruption safety and valid-signal metrics by configuration.}
\label{tab:results_synthetic_safety}
\footnotesize
\begin{tabularx}{\textwidth}{l Y r r r}
\toprule
\textbf{Config.} & \textbf{Configuration} & \textbf{Unsafe repair} & \textbf{Unnecessary modification} & \textbf{Valid-signal preservation} \\
\midrule
A0 & Schema-only & 0.000 & 0.222 & 0.000 \\
A1 & Profiling/checks & 0.064 & 0.167 & 1.000 \\
A2 & LLM + profiling & 0.000 & 0.111 & 0.556 \\
A3 & Python tools & 0.000 & 0.000 & 0.000 \\
A4 & Python + controlled evidence & 0.000 & 0.000 & 0.000 \\
A5 & Source-ranked evidence & 0.000 & 0.000 & 0.000 \\
A6 & Full conservative & 0.000 & 0.000 & 0.000 \\
\bottomrule
\end{tabularx}
\end{table}

A zero valid-signal preservation rate should be interpreted with care because the metric is defined only over injected valid-signal cases that were detected as issues. A value of zero may therefore also occur when no injected valid-signal case was detected.

The repair and safety metrics require joint interpretation. A1 achieved the highest safe-repair rate at 0.770 and the highest valid-signal preservation rate at 1.000, but it also produced unsafe repairs and unnecessary modifications. The high safe-repair rate does not mean that the proposed corrections were accurate: its repair-match rate was 0.000. In fact, only A3 achieved a non-zero repair-match rate, at 0.028, showing that successful direct correction remained very limited across all configurations.

A6 produced no unsafe or unnecessary modifications in the synthetic summary, but it also produced no repair-class decisions and therefore had zero safe repair and zero repair match. Its behavior is better interpreted as conservative non-repair rather than successful direct repair. These results show a clear trade-off between attempting repair and avoiding inappropriate modification.

At dataset level, averaged over configurations and repetitions, EHR achieved the highest synthetic detection F1-score at 0.405, followed by German Credit at 0.380 and Radiation at 0.280. These differences are consistent with H5, showing that the relative effectiveness of the evaluated configurations varies across datasets. The current experiment does not isolate which specific dataset characteristics caused these differences.

Original-data results are shown in Tables~\ref{tab:results_natural_behavior} and~\ref{tab:results_natural_operational}. These values describe agent behavior, not accuracy. A1 produced the largest mean number of issues and decisions, 10.000 for both, while A5 produced the fewest decisions, 2.222. A6 had evidence coverage of 0.556 and a human-review rate of 0.111 in the original-data summary, but it also had the highest original-data runtime and token consumption.

\begin{table}[!htbp]
\centering
\caption{Original-data descriptive behavior and evidence indicators by configuration. Values are means over nine original-data runs per configuration: three datasets and three repeated runs.}
\label{tab:results_natural_behavior}
\scriptsize
\setlength{\tabcolsep}{4pt}
\resizebox{\textwidth}{!}{%
\begin{tabular}{l l r r r r r r}
\toprule
\textbf{Config.} & \textbf{Configuration} & \textbf{Issues} & \textbf{Decisions} & \textbf{Evidence cov.} & \textbf{Unsupported rules} & \textbf{Conserv. flag} & \textbf{Human review} \\
\midrule
A0 & Schema-only & 3.778 & 3.778 & 0.089 & 0.667 & 0.653 & 0.093 \\
A1 & Profiling/checks & 10.000 & 10.000 & 0.627 & 0.040 & 0.453 & 0.000 \\
A2 & LLM + profiling & 5.222 & 2.667 & 0.236 & 0.167 & 0.444 & 0.111 \\
A3 & Python tools & 3.556 & 3.556 & 0.548 & 0.074 & 0.644 & 0.000 \\
A4 & Python + controlled evidence & 6.444 & 3.111 & 0.469 & 0.028 & 0.542 & 0.000 \\
A5 & Source-ranked evidence & 4.000 & 2.222 & 0.356 & 0.111 & 0.444 & 0.022 \\
A6 & Full conservative & 3.778 & 3.000 & 0.556 & 0.037 & 0.556 & 0.111 \\
\bottomrule
\end{tabular}%
}
\end{table}

\begin{table}[!htbp]
\centering
\caption{Original-data operational cost and structured-output repair by configuration. Values are means over nine original-data runs per configuration.}
\label{tab:results_natural_operational}
\footnotesize
\begin{tabularx}{\textwidth}{l Y r r r}
\toprule
\textbf{Config.} & \textbf{Configuration} & \textbf{Runtime (s)} & \textbf{Total tokens} & \textbf{JSON repair} \\
\midrule
A0 & Schema-only & 25.209 & 4841.444 & 0.222 \\
A1 & Profiling/checks & 0.081 & 0.000 & 0.000 \\
A2 & LLM + profiling & 52.616 & 9346.222 & 0.556 \\
A3 & Python tools & 61.958 & 9392.444 & 0.111 \\
A4 & Python + controlled evidence & 43.521 & 9918.111 & 0.333 \\
A5 & Source-ranked evidence & 34.356 & 10132.889 & 0.222 \\
A6 & Full conservative & 138.938 & 14803.000 & 0.333 \\
\bottomrule
\end{tabularx}
\end{table}

Evidence metrics show partial but limited grounding. In synthetic runs, evidence coverage was 0.870 for A1, 0.778 for A4, and 0.722 for A2. A5 and A6 had lower evidence coverage, at 0.630 and 0.519, but achieved the lowest unsupported-rule rates, both at 0.037. Citation alignment was non-zero only for the configurations with citation-alignment checks enabled, reaching 0.036 for A5 and 0.041 for A6. These low values show that retrieving, ranking, and recording evidence did not result in strong decision-level citation alignment. The results therefore distinguish evidence availability and rule support from direct alignment between cited evidence and individual cleaning decisions.

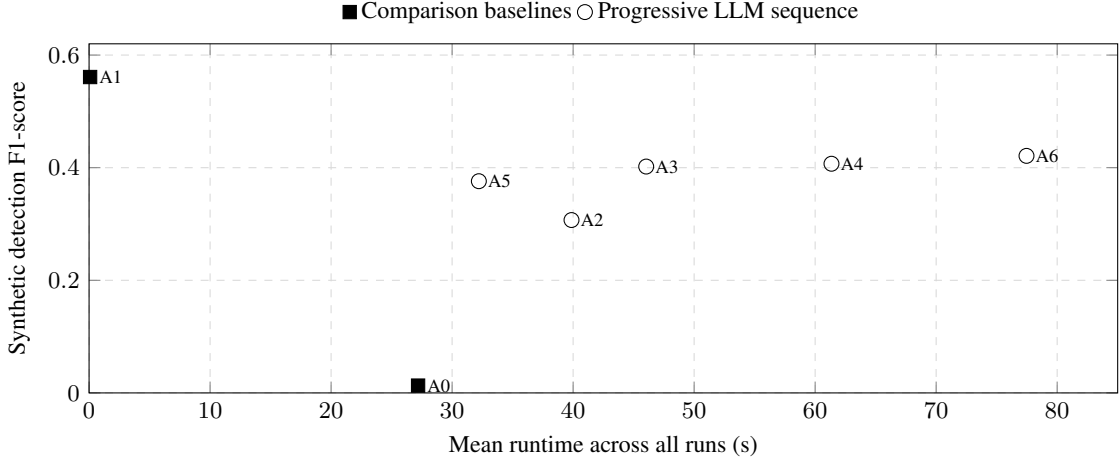
\begin{figure}[!htbp]
\centering
\begin{tikzpicture}
\begin{axis}[
    width=0.92\textwidth,
    height=6.2cm,
    xmin=0,
    xmax=85,
    ymin=0,
    ymax=0.62,
    xlabel={Mean runtime across all runs (s)},
    ylabel={Synthetic detection F1-score},
    xmajorgrids=true,
    ymajorgrids=true,
    grid style={dashed,gray!30},
    axis line style={black},
    tick label style={font=\small},
    label style={font=\small},
    legend style={
        font=\small,
        draw=none,
        at={(0.5,1.03)},
        anchor=south,
        legend columns=2
    },
]

\addplot+[
    only marks,
    mark=square*,
    mark size=2.5pt,
    black,
    mark options={fill=black,draw=black},
    nodes near coords,
    point meta=explicit symbolic,
    nodes near coords style={font=\scriptsize, anchor=west}
] table[meta=label] {
runtime f1 label
27.189 0.013 A0
0.086  0.561 A1
};
\addlegendentry{Comparison baselines}

\addplot+[
    only marks,
    mark=o,
    mark size=2.8pt,
    black,
    mark options={draw=black},
    nodes near coords,
    point meta=explicit symbolic,
    nodes near coords style={font=\scriptsize, anchor=west}
] table[meta=label] {
runtime f1 label
39.867 0.307 A2
46.039 0.402 A3
61.350 0.407 A4
32.210 0.376 A5
77.466 0.421 A6
};
\addlegendentry{Progressive LLM sequence}

\end{axis}
\end{tikzpicture}
\caption{Configuration-level trade-off between synthetic detection performance and mean runtime across all 18 runs per configuration. Filled squares denote the comparison baselines, while open circles denote the progressive LLM-based sequence.}
\label{fig:f1_runtime_tradeoff}
\end{figure}

Operational cost varied across configurations. A1 was the fastest because it used deterministic profiling and no LLM calls. Among the LLM-based configurations, token use generally increased as additional capabilities were enabled, while runtime did not increase monotonically. Across all runs, A6 had the highest mean runtime and token use, at 77.466 seconds and 12916.222 tokens. In original-data runs, A6 reached 138.938 seconds and 14803.000 tokens. Reproducibility, measured as exact agreement of decision signatures, was highest for A1 at 0.500, followed by A5 and A6 at 0.444. Because this metric requires exact agreement of issue identifiers and decision classes, it may count semantically similar outputs as different. Figure~\ref{fig:f1_runtime_tradeoff} visualises the configuration-level trade-off between synthetic detection performance and mean runtime across all runs.

Four runs required a second attempt before success, but all completed successfully. Structured-output enforcement was needed for consistent metric computation, as reflected by the JSON repair and postprocessing counts in Table~\ref{tab:results_execution_audit}. No separate qualitative case analysis was performed, so qualitative interpretation is limited to patterns recorded in the saved run artifacts and metrics.

\section{Discussion}
\label{sec:discussion}

The results highlight a central characteristic of reference-free agentic data cleaning: adding capabilities changes different aspects of system behaviour, but does not lead to a monotonic improvement in overall cleaning performance. The strongest synthetic detection result was obtained by A1, the deterministic profiling-and-checking baseline, rather than by the most complete LLM configuration. At the same time, the schema-only LLM baseline A0 performed poorly, while incorporating profiling information in A2 substantially improved detection behaviour. This suggests that structured, data-derived evidence plays an important role in effective cleaning decisions and that LLM reasoning alone is not sufficient. However, because A0 and A1 differ in both reasoning mechanism and available capabilities, their difference should not be interpreted as a single-factor causal effect.

Within the progressive LLM sequence, the clearest capability-level improvement appeared when executable tools were introduced. Moving from A2 to A3 increased synthetic detection F1-score from 0.307 to 0.402, and the generated reversible scripts were successfully executed. Even so, the repair-match rate remained only 0.028. Tool access therefore improved the agent's ability to validate suspected problems without resolving the more difficult question of what the correct replacement value should be. This distinction is important in reference-free cleaning because identifying that a value may be problematic and determining how it should be repaired are separate tasks that require different levels of evidence. These observations are directionally consistent with H1 and help answer RQ1 and RQ2.

The evidence-related configurations show a similar distinction between having evidence available and actually grounding decisions in that evidence. A5 and A6 achieved the lowest synthetic unsupported-rule rates, both at 0.037, suggesting that source-ranking and citation controls were associated with better-supported validation rules. Their decision-level citation-alignment scores, however, remained very low at 0.036 and 0.041. The controlled local corpus improved reproducibility by keeping the available evidence fixed across runs, but retrieval, ranking, and provenance logging did not ensure that the evidence cited for a decision directly supported that decision. Evidence retrieval should therefore not be treated as equivalent to evidence grounding. In its current form, the system is better characterised as evidence-controlled and provenance-aware than as a mature citation-grounded autonomous repair system. Stronger decision-level alignment and independent validation of supporting evidence would be needed before such mechanisms could justify autonomous high-impact repairs. These findings address RQ3 and are consistent with H2.

The conservative policy introduced in A6 produced another important trade-off. A6 made no unsafe or unnecessary modifications in the synthetic evaluation, but A5 had already achieved zero on both indicators before the conservative policy was added. The A5--A6 comparison therefore does not provide clear support for H3 on these two safety measures. The more noticeable effect of A6 was behavioural: it shifted the system toward abstention and escalation, producing no repair-class decisions in the synthetic setting and referring some original-data cases for human review. This behaviour is broadly consistent with H4 and addresses RQ4, but it is better interpreted as conservative non-repair than as improved repair effectiveness. In applications where an incorrect modification may be more harmful than leaving an anomaly unresolved, this type of behaviour may still be desirable.

These additional controls also came with operational costs. A6 had the highest overall runtime and token consumption, whereas the deterministic A1 baseline was considerably less expensive because it required no LLM inference. Runtime did not increase monotonically across the intermediate LLM configurations, indicating that cost depends not only on how many capabilities are enabled but also on inference behaviour, retrieval, and tool execution. The results for RQ5 reinforce the multi-objective nature of reference-free cleaning: detection effectiveness, repair accuracy, evidence support, safety, reproducibility, and operational cost cannot be optimised independently. Simpler configurations may be sufficient for exploratory detection, whereas stronger evidence controls and conservative decision policies may be more appropriate when the consequences of an incorrect repair are greater.

The dataset-level results also suggest that configuration effectiveness depends on data characteristics and cleaning context. Mean synthetic detection F1-score was highest for EHR at 0.405, followed by German Credit at 0.380 and Radiation at 0.280, which is directionally consistent with H5. The experiment does not isolate the causes of these differences. Temporal patterns and rare events, for example, may contribute to the lower Radiation results, while the prompt sample of up to 10 records may provide insufficient information for longer temporal patterns, rare environmental events, or entity-level relationships. These explanations remain hypotheses and would require targeted experiments before they could be attributed to specific dataset properties.

The original-data runs provide a complementary view of system behaviour, but they should not be interpreted as an accuracy evaluation. Without a trusted clean reference or independent adjudication, these runs show how the configurations detect issues, use evidence, make conservative decisions, escalate cases, and consume resources rather than whether the resulting decisions are objectively correct. Keeping this distinction between controlled synthetic evaluation and original-data behavioural auditing is important for avoiding unsupported accuracy claims in a reference-free setting.

Taken together, the hypotheses receive different levels of directional support. The A2--A3 improvement is consistent with the executable-validation component of H1, while the lower unsupported-rule rates of the source-ranked configurations, together with weak citation alignment, are consistent with H2. H3 is not clearly supported by the A5--A6 comparison because A5 already had zero unsafe and unnecessary modifications before the conservative policy was introduced. H4 is broadly consistent with the more conservative behaviour of A6 and its associated trade-offs in direct repair and operational cost. Finally, the differences across datasets are consistent with H5, although the specific dataset characteristics responsible for those differences were not isolated.

Overall, the findings suggest that agentic data-cleaning systems are better viewed as combinations of separately evaluable capabilities than as monolithic agents whose performance is expected to improve simply by adding more reasoning, retrieval, or tools. Profiling, executable validation, evidence controls, conservative decision policies, and abstention affected different aspects of behaviour, and no configuration dominated across detection, repair, safety, evidence grounding, reproducibility, and operational cost. Capability composition should therefore be selected according to the cleaning objective, the evidence available, and the consequences of an incorrect modification. The main contribution of this study is an auditable framework and evaluation methodology for examining these trade-offs, rather than a claim that an autonomous agent can safely clean arbitrary datasets without expert oversight.

\section{Threats to Validity}
\label{sec:threats}

The findings should be interpreted in light of several threats to validity that reflect the exploratory ablation design, the reference-free setting, and the controlled implementation choices used in the experiment.

\begin{itemize}

\item \textbf{Construct validity.} The metrics are operational approximations of safe reference-free cleaning and do not capture every downstream use of a cleaned dataset. Synthetic precision, recall, F1-score, expected-repair match, and valid-signal preservation are meaningful only when injected ground truth is available. Original-data metrics describe behavior, evidence use, cost, and reproducibility, not accuracy. Safe repair is also context-dependent: a transformation may be acceptable for reporting but unsafe for anomaly detection, auditing, or forensic analysis. In addition, the individual knowledge categories defined in the taxonomy were not independently ablated, so the experiment cannot determine whether any single category is necessary or sufficient. The structured context $\Gamma_X$ helps represent these factors but cannot replace domain judgement.

\item \textbf{Evidence and citation validity.} Evidence retrieval used a controlled local corpus rather than live online retrieval. Evidence-strength and citation-alignment scores therefore reflect the implemented evidence records and automated scoring procedures, not independent human verification of source quality or decision support. The low citation-alignment values, and the fact that many final decisions cited profiling or executable-check evidence, limit claims about external evidence grounding.

\item \textbf{Internal validity.} Synthetic corruption provides partial ground truth, but injected errors may be more regular, visible, or reversible than natural errors. Original-data outputs were not manually adjudicated, so they cannot support detection or repair accuracy claims. A0 and A1 are contrasting baselines rather than a single-factor ablation because they differ in both reasoning mechanism and available profiling capabilities. Component-level interpretation is therefore stronger for the progressive A2--A6 comparisons. Prompt wording, output schema, temperature, model version, provider behavior, and postprocessing rules may also affect LLM-based decisions. Fixed prompts, three repeated runs, saved artifacts, and postprocessed metrics reduce these effects but do not remove them. Three repetitions are also insufficient for strong inferential claims about run-to-run variability.

\item \textbf{External validity.} The experiment uses three datasets from finance, healthcare, and environmental monitoring, but the results may not generalise to legal, geospatial, multilingual, relational, streaming, or image-derived data. The sample of up to 10 records supplied to the LLM may be insufficient for long temporal patterns, rare events, or entity-level constraints. The study also used one LLM and provider setting for all LLM-based runs, so generalisation across models and providers remains to be evaluated.

\item \textbf{Reproducibility and temporal validity.} The controlled local evidence corpus improves reproducibility compared with live online retrieval but limits the range and freshness of available evidence. LLM services may also change because of model updates, routing, moderation, or infrastructure changes. The experiment records the model configuration, provider, inference parameters, retry behavior, responses, run metadata, and postprocessed metrics needed to audit the reported runs. Future experiments using online retrieval should preserve source snapshots or archived documents, while locally hosted or pinned open-weight models could further reduce model-version variability.

\item \textbf{Implementation and evaluation bias.} Profiling scripts, corruption generators, source-ranking functions, citation-alignment checks, reversible-script generation, JSON repair, postprocessing, and metric computations may contain bugs or hidden assumptions. Taxonomy definitions, corruption scenarios, conservative policies, evaluation metrics, and evidence-ranking weights also reflect researcher judgement. Saved intermediate artifacts, execution logs, output audits, and explicit scoring logic reduce this risk but do not eliminate it. Independent domain experts and multiple reviewers should evaluate the framework in future replications.

\end{itemize}

\section{Conclusion}
\label{sec:conclusion}

This study examined how different capabilities affect reference-free data cleaning with LLM agents. It proposed an evidence-grounded framework that combines structured context, profiling, LLM reasoning, executable checks, controlled local evidence retrieval, source ranking, citation alignment, conservative repair, reversible scripts, and provenance logging. The experiment produced 126 successful runs across three datasets, two evaluation settings, and seven configurations. By separating controlled synthetic-corruption evaluation from original-data descriptive auditing, the study avoids treating unverified anomalies in the original data as ground-truth errors.

The results show that the evaluated capabilities affect different aspects of the cleaning process and do not produce consistent improvements across all metrics. The deterministic profiling baseline achieved the highest synthetic detection F1-score, while adding executable tools improved detection within the progressive LLM-based sequence. Source-ranking and citation controls were associated with low unsupported-rule rates, although citation alignment remained weak. The full conservative configuration avoided unsafe and unnecessary modifications in the synthetic summary, but these values were already zero in A5, and A6 performed no successful direct repairs. It also introduced the highest overall runtime and token use. Overall, no configuration dominated across detection, repair, safety, evidence grounding, reproducibility, and operational cost. Reference-free agentic data cleaning should therefore be viewed as a multi-objective problem involving trade-offs among these criteria.

Future work should evaluate additional datasets, models, and evidence providers; compare controlled local and online evidence retrieval; include independent domain-expert review of original-data outputs; improve decision-level evidence alignment, repair evaluation, and semantic reproducibility measures; and examine individual knowledge categories through controlled ablation. In particular, independently adding or removing schema, temporal, entity, domain, policy, task, and risk knowledge would help determine which forms of knowledge are useful for different classes of reference-free cleaning decisions.

\section*{Declarations}

\subsection*{Conflict of Interest}
The author declares that there are no conflicts of interest.

\subsection*{Ethics, Informed Consent, and Data Use}
This study did not recruit human participants or collect new human-subject data. The experiments used publicly available datasets obtained from the sources cited in the manuscript. No additional informed consent was sought for this secondary analysis.

\subsection*{Use of AI Technology}
AI-based tools, including ChatGPT, were used to assist with language editing, clarity, manuscript organisation, and consistency checking. The author independently reviewed and verified all scientific claims, analyses, interpretations, and conclusions and remains fully responsible for the manuscript content. AI tools were not used to execute the reported experiments or generate the experimental data or results.

\section*{Data and Code Availability}
The datasets used in this study are publicly available from the sources cited in the manuscript. A blinded repository containing the source code, experiment scripts, configuration files, controlled evidence corpus, and saved experimental artefacts used to produce and inspect the reported results is available for review at: \url{https://drive.google.com/drive/folders/1Cqfxg_BxLBh_eMnMsFyTDPSQuSmo2uTW?usp=sharing}. The blinded repository will be replaced by a permanent public archival record after acceptance.

\bibliographystyle{unsrt}
\bibliography{Ref}

\end{document}